\documentclass[letterpaper, 10 pt, conference]{ieeeconf}  

\IEEEoverridecommandlockouts                              

\usepackage{graphics} 
\usepackage{graphicx}
\usepackage{times} 
\usepackage{amsmath} 
\usepackage{cite}
\usepackage{tabularx}
\usepackage{booktabs}
\usepackage{multirow}

\title{\LARGE \bf
Real-time 3D semantic occupancy prediction for autonomous vehicles using memory-efficient sparse convolution
}

\author{Samuel Sze and Lars Kunze$^{1}$
\thanks{$^{1}$Samuel Sze and Lars Kunze is with the Cognitive Robotics Group, Oxford Robotics Institute, Department of Engineering Science, University of Oxford: \texttt{\small (samuels,lars)@robots.ox.ac.uk}}%
}

\begin{document}

\maketitle
\thispagestyle{empty}
\pagestyle{empty}

\begin{abstract}

In autonomous vehicles, understanding the surrounding 3D environment of the ego vehicle in real-time is essential. A compact way to represent scenes while encoding geometric distances and semantic object information is via 3D semantic occupancy maps. State of the art 3D mapping methods leverage transformers with cross-attention mechanisms to elevate 2D vision-centric camera features into the 3D domain. However, these methods encounter significant challenges in real-time applications due to their high computational demands during inference. This limitation is particularly problematic in autonomous vehicles, where GPU resources must be shared with other tasks such as localization and planning. In this paper, we introduce an approach that extracts features from front-view 2D camera images and LiDAR scans, then employs a sparse convolution network (Minkowski Engine), for 3D semantic occupancy prediction. Given that outdoor scenes in autonomous driving scenarios are inherently sparse, the utilization of sparse convolution is particularly apt. By jointly solving the problems of 3D scene completion of sparse scenes and 3D semantic segmentation, we provide a more efficient learning framework suitable for real-time applications in autonomous vehicles. We also demonstrate competitive accuracy on the nuScenes dataset.

\end{abstract}

\section{INTRODUCTION}

The rise of Autonomous Vehicles (AV) heralds a transformative era in transportation, moving us to potentially more efficient and safer alternatives. At the heart of these autonomous systems is a software stack composed of perception, localization, planning, and control. Among these, perception is crucial as it enables the vehicle to interpret and understand its surroundings, serving as the foundation for all subsequent vehicle decision-making. Typically, AV's deploy a suite of sensors such as cameras and LiDARs to perceive their surroundings. Cameras operate in 2D perspective image space, capturing essential color and texture details but lacking 3D spatial information. In contrast, sensors like LiDAR inherently provide 3D coordinates, albeit often with lower spatial resolution. For a holistic 3D scene understanding, both pixel-level information from cameras and sparse spatial information from LiDARs should be leveraged.

\begin{figure}[thpb]
    \centering
    \parbox{3.3in}{%
        \centering
        \includegraphics[scale=0.48]{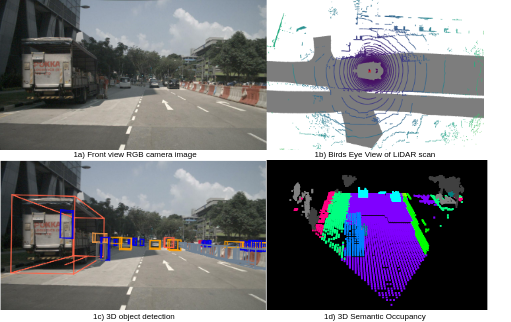} 
    }
    \caption{Scene understanding methods using camera and LiDAR in nuScenes. Top two images shows perceived sensor output. Bottom two images shows scene predictions made from sensor output.}
    \label{intro}
\end{figure}

Traditional scene understanding uses 3D object detection methods that take in sensor information from cameras and LiDARs and project 3D bounding boxes across objects of interest such as vehicles, pedestrians, and traffic lights. However, while 3D object detection provides information about object instances, it often fails to capture complex object geometries and background information such as drivable space and buildings \cite{occ3d, openoccupancy}. An alternative approach to scene understanding is constructing 3D semantic occupancy maps, which partition the environment into a structured grid map with a predefined resolution, allowing each grid cell to be assigned a semantic label.

Nevertheless, 3D semantic occupancy prediction of outdoor scenes proves to be difficult due to the scenes' sparsity. Large areas such as the sky and occluded space captured by the camera often contain no relevant information. Most autonomous vehicles use rotary beam LiDARs which are inherently sparse, leading to overall sparse LiDAR data. To accurately interpret and fill in the missing parts of the environment, these algorithms must incorporate elements of 3D scene completion in addition to 3D semantic segmentation.

In this work, we focus on applying the existing sparse 3D convolution engine, Minkowski Engine \cite{choy20194d}, to a fused sensor representation of 2D camera and 3D LiDAR data for 3D semantic occupancy prediction. While the Minkowski Engine itself is a mature method, our key contribution lies in the innovative application of this sparse convolution framework to the problem of 3D semantic occupancy prediction. The fused sensor data enables sparse convolution to effectively handle outdoor scenes, allowing for accurate scene completion and semantic segmentation. Thereby, our main contributions are listed as the following: 
\begin{itemize}
    \item We propose the design of a novel sparse 3D convolution (Minkowski Engine \cite{choy20194d}) model to perform 3D semantic occupancy prediction that jointly solves the problem of scene completion and semantic segmentation.
    \item We evaluate the model's 3D scene completion and semantic segmentation performance, achieving competitive accuracy against other algorithms on nuScenes dataset \cite{occ3d}.
    \item We conduct time and memory usage evaluations to ensure model's real-time inference capabilities are close to human perception rates of 20 - 30 frames per second (FPS).
\end{itemize}

\section{Related Work}

\subsection{Camera View Transformation}
Camera view transformation techniques can be broadly bifurcated into two categories: extending 2D camera representations into a 3D domain or initializing a 3D domain and distilling 2D information onto it. The former aims to "lift" the camera image from perspective view to a 3D view, whereas the latter aims to "pull" camera information onto a predefined 3D space. Past lifting techniques includes probabilistic depth predictions for each camera pixel \cite{lss,caddn, bevfusion, fiery, learningtolookaroundobjects}, as well as Multi-layer Perceptron (MLP) methods \cite{pyroccnet, fishingnet}. A current popular pulling technique is using cross-attention based transformers where queries relayed from a discretized 3D space draws out keys and queries from a 2D perspective image\cite{bevformer, persformer, image2map, voxformer, scpnet}. 

Simple-BEV \cite{simplebev} suggests that the accuracy of Bird's Eye View (BEV) semantic prediction and similarly, 3D semantic occupancy prediction, often hinges more on factors like batch size, image resolution, and sensor fusion rather than sophisticated view transformation algorithms. Moreover, it is important to acknowledge that transforming views from 2D to 3D is inherently an ill-posed problem due to the absence of depth information in 2D images. This limitation highlights the necessity of employing deterministic methods that rely on additional sensors such as LiDARs. Arguing against the use of tools like LiDAR due to their cost overlooks their essential role in providing a fast and accurate method for view transformation. 

\subsection{3D Semantic Scene Completion}
Developments in 3D semantic scene completion of outdoor self-driving scenes is mainly spurred by well-curated datasets. With Semantic-KITTI \cite{semantickitti}, many LiDAR-based semantic scene completion methods appeared \cite{s3cnet, sscnet, SATnet}. Given LiDAR as the only input, many of these methods do not require view transformation, simplifying the problem. However, most mainstream methods are built upon indoor LiDAR dataset such as NYUv2 \cite{nyuv2} and ScanNet\cite{scannet}, and do not translate well to outdoor scenarios due to the sparsity of LiDAR points. Specific methods built upon Semantic-KITTI employed innovative solutions to densify the 3D scene. Notably, S3CNet \cite{s3cnet} also uses Sparse convolution to balance computational budget while achieving state-of-the-art results. Nevertheless, LiDAR data lacks color and texture information, limiting its ability to discern objects with similar shapes but different visual features, such as distinguishing a street pole from a traffic light. Hence, LiDAR-only methods still require some level of feature engineering such as spherical projection to a range image, creating truncated signed distance function (TSDF) volumes, and estimating surface normal. These feature engineering inevitably requires additional tuning while introducing some level of noise.

More recently, multi-modal datasets such as nuScenes \cite{nuscenes} and its extension, Occ3D-nuScenes \cite{occ3d}, have enabled the development of vision-centric 3D semantic scene completion, primarily using cameras as the model input. Vision-centric methods, initially explored by Tesla \cite{tesla} and further researched in various research studies \cite{bevformer, monoscene, surroundocc, tpvformer}, employs cross-attention to geometrically relate 2D camera features into a 3D voxel grid, which is then used for BEV predictions or 3D semantic occupancy predictions. Efforts like deformable attention \cite{voxformer} and coarse-to-fine strategies \cite{occ3d} have been implemented to reduce the computational burden. Yet, a significant downside remains in the inherent dimensionality of establishing a dense 3D voxel grid, where attention mechanisms and transformers struggle to manage efficiently. Additionally, relying solely on camera data requires the model to simultaneously address the task of identifying and denoising distorted occluded areas on top of scene completion and semantic segmentation, posing a complex and challenging training task.


\subsection{Sparse Convolution}
For 3D and higher dimension data, it is often inefficient to parse such data through traditional dense 3D convolutions as most grid cells are empty to begin with. In the context of 3D semantic occupancy prediction, sparse convolution operates on spatially sparse 3D data. It only considers 3D points that are specified, often ones which contains LiDAR or camera information, whilst discarding meaningless empty 3D points. Sparse Convolution Networks, like Minkowski Engine \cite{choy20194d}, have been effectively utilized in indoor scenes \cite{minkindoor} and for single object analysis \cite{minkobject}. However, their application in outdoor driving scenes, particularly with the availability of multi-modal outdoor datasets, remains an area for exploration. Given their proficiency in selectively processing meaningful 3D points, these networks are well-suited for the large and sparse areas typical of LiDAR and camera data capturing outdoor environments. Adapting sparse convolution networks to outdoor 3D semantic occupancy prediction presents a promising avenue for addressing the computational challenges currently faced in this field.

\begin{figure*}[thpb]
    \centering
    \parbox{6in}{
        \centering
        \includegraphics[scale=0.7]{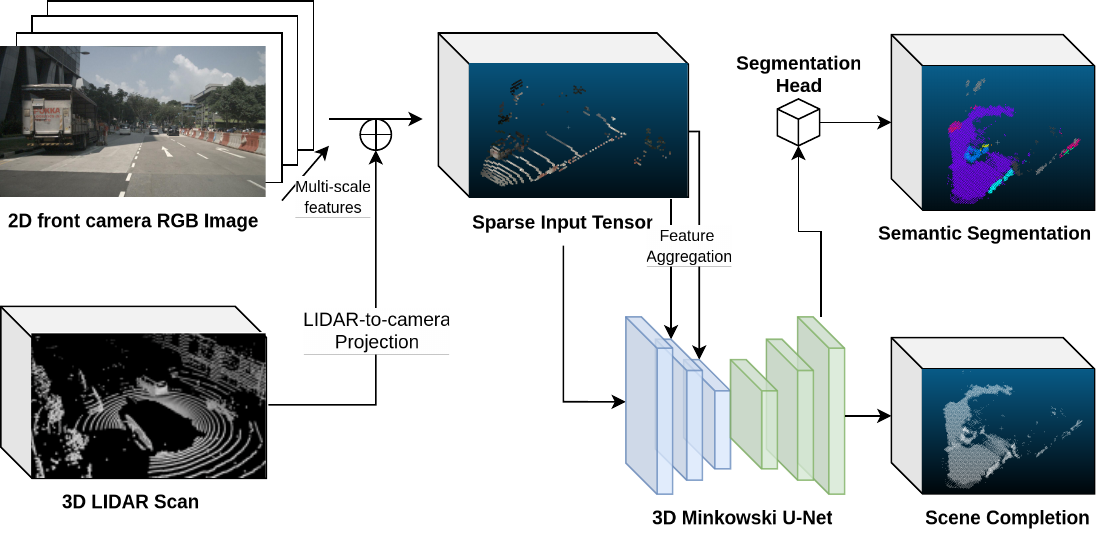}
    }
    \caption{Overview of System Pipeline}
    \label{system_arch}
\end{figure*}

\section{Problem Formulation} 
Formally, we define the 3D semantic occupancy prediction within the context of one front-view camera positioned to view directly ahead of an autonomous vehicle. Given a 2D camera image of dimensions $W \times H$, where each pixel $i_{ij}$ contains RGB values $(R, G, B)$ 
, the objective is to use view transformation to project, and sparse convolution to densify and segment image features into a 3D voxel space $V$ with dimensions $X \times Y \times Z$ in real-time. We define real-time to be at least 20 frames per second (20 FPS), with the goal of reaching 30 FPS to match human perception rate. Each voxel $v_{xyz}$ in this 3D space should encapsulate semantic and spatial information based on camera and LiDAR. The expected outcome is a labeled 3D voxel grid, where each voxel is affiliated with one of $L$ class labels, denoted as $L = \{l_1, l_2, ..., l_C\}$, with $C$ representing the aggregate number of distinct classes. 

\section{Method}
We utilize the Occ3D-nuScenes dataset \cite{occ3d}, an extension of the nuScenes dataset, which is designed for 3D semantic occupancy prediction and benchmarking. It comprises of 28,130 Front camera RGB training images and 6019 validation images, as well as their corresponding voxelized ground truth semantic occupancy grid. The input LIDAR scans are obtained from the original nuScenes \cite{nuscenes} dataset. 

\subsection{System Pipeline}
The system pipeline is shown in Figure \ref{system_arch}. Using a calibrated extrinsic and intrinsic matrix, we first project LIDAR points onto the RGB image space. Each projected point is fused with RGB color information through bilinear interpolation. Subsequently, the fused data is unprojected back into the 3D space. We also perform feature extraction using EfficientNetV2 \cite{efficientnetv2} at layer 3,4 and 6 to extract higher level 2D camera features and lift them to the 3D space using the same projection method but scaled accordingly to accommodate for a reduction in spatial dimensions. The deterministic nature of the LiDAR-to-camera projection, assuming the sensors are adeptly calibrated, offers an immediate and precise depth-location correspondence - a key attribute for real-time tasks in autonomous vehicles.

Initial representation now has attributes of spatial coordinates (\( x, y, z \)) and feature values (\([feats] +  LiDAR intensity\)). Spatial coordinates is voxelized by a resolution of 0.4m, discretizing the pointcloud into a 3D tensor, \( T_{\text{sparse}} \). \( T_{\text{sparse}} \) has dimensions of 200 x 200 x 16 grid units in tensor shape, covering a physical boundary of [-40m, -40m, -1m] to [40m, 40m, 5.4m]. \( T_{\text{sparse}} \) is later converted to Coordinate List format (COO) with only non-empty grids included during sparse convolution operations. We leverage the Minkowski Engine generative completion algorithm \cite{generative_me} to densify the sparse tensor: \(T_{\text{dense}} = \mathcal{M}(T_{\text{sparse}})\), where \( \mathcal{M} \) represents the generative scene completion operation. Scene completion is guided by Occ3D-nuScenes \cite{occ3d} dataset ground truth labels with the grid resolution to match \( T_{\text{sparse}} \)'s dimension.

We carry out semantic segmentation on \( T_{\text{dense}} \), directly supervised by ground truth semantic labels. Both the semantic segmentation and scene completion step is trained as a multi-task problem, with the semantic segmentation having an additional network of smaller size, predicting outputs over 18 semantic categories. The final 3D semantic occupancy prediction has 200 x 200 x 16 voxel grid with 0.4m resolution.

\subsection{Minkowski Engine}
 Minkowski Engine processes sparse, high-dimensional data by focusing on non-zero data points and allowing adaptability in the kernel's shape and size. In other words, it follows a generalized convolution \cite{choy20194d}.  First, a sparse tensor is defined as a combination of a hash-table of coordinates and their corresponding features. Specifically, this is denoted as \( u = [C_{n \times d}, x_{n \times m}] \), where \( C_{n \times d} \) represents the coordinates in COO format and \(x_{n \times m} \) corresponds to the feature vectors. The convolution operation within this framework is given by:

\begin{equation}
x_{\text{out}}^u = \sum_{i \in \mathcal{N}^D(u, C_{\text{in}})} W_i \cdot x_{\text{in}}^{u + i} \quad \text{for} \quad u \in C_{\text{out}}
\label{mink_equation}
\end{equation}

In equation \ref{mink_equation}, \( x_{\text{out}}^u \) is the output feature at point \( u \). \( \mathcal{N}^D(u, C_{\text{in}}) \) denotes the set of offsets defining the neighborhood of input coordinates around \( u \) that plays a part in the sparse convolution layer.  \( W_i \) are the learnable weights of the convolution kernel, and \( x_{\text{in}}^{u + i} \) is the input feature at the position offset by \( i \) from \( u \). 


\begin{figure*}[thpb]
    \centering
    \parbox{7in}{
        \centering
        \includegraphics[scale=0.7]{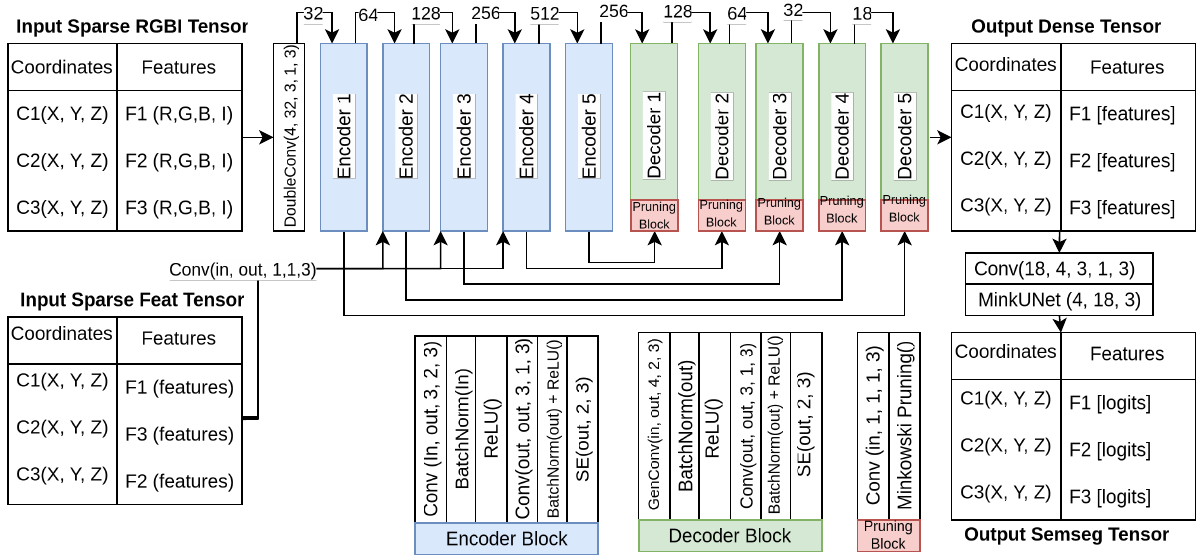}
    }
    \caption{Minkowski Engine Sparse Convolution Network Architecture. Scene Completion U-Net is shown in detail. MinkUNet is the Semantic Segmentation U-Net.}
    \label{network_arch}
\end{figure*}

\subsection{Network Architecture}
With reference to figure \ref{network_arch}, the proposed 3D semantic occupancy prediction model utilizes 2 U-Net-like \cite{unet} like encoder-decoder networks sequentially. Both U-Net shares similar structure, with the scene completion U-Net having deeper layers and specialized modules such as Squeeze and Excite \cite{selayer}, Generative Convolution Transpose \cite{generative_me} and Sparse Pruning. An encoder block contains a double convolution Layer and a squeeze and excite (SE) layer. The convolution layer features batch normalization (BN) and ReLU activation to standardize the variable density of sparse data.
The SE layer dynamically recalibrates channel-wise features through global average pooling, enhancing the model's focus on key features across the input sparse data. For certain encoder blocks, image features extracted from EfficientNetV2 \cite{efficientnetv2} are also concatenated before further downsampling.

The decoder module for scene completion is designed for generative tasks. The process begins with generative transposed convolution capable of generating new coordinates with the outer product of the weight kernel and the input coordinates. A SE layer is attached after generation. At each decoder layer, a skip connection from the corresponding encoder is also added. Following that, a 3D binary classification convolution is applied to the sparse tensor which assigns a decision value to prune. The decision threshold is trainable with ground truth supervision. Redundant features are pruned using Minkowski Engine's built in pruning module. Overall, the decoder generates, upsamples, classifies, and then prunes voxels to iteratively create accurate 3D representation at each layer depth.

Upon reaching the final decoder layer, an additional semantic segmentation U-Net head is attached for the sole purpose of semantic segmentation of the densified sparse tensor. This U-Net features sparse convolution layers and skip connections, with bottleneck channel length of 256. 

\subsection{Loss Function}
A multi-task loss function is defined to train the 3D Semantic Occupancy prediction model. The two terms are balanced by a Lambda ($\lambda$) constant empirically set to 0.5. Experiments conducted on different $\lambda$ constants is represented in the ablation study.  

\begin{equation}
    \mathcal{L}_{\text{loss}} = \mathcal{L}_{\text{completion}} + \lambda \mathcal{L}_{\text{segmentation}}
\end{equation}

\(\mathcal{L}_{\text{completion}}\) is a voxel based binary cross entropy with logits loss (\(\sum_{i,j,k} \lambda_{\text{BCE}}(p_{ijk}, y_{ijk})\)). It is computed over the volumetric occupancy on each decoder layer. This loss also tunes the decision threshold used for pruning generated voxels.

\begin{equation}
\mathcal{L}_{\text{segmentation}}(z, y) = - \frac{1 - \beta}{1 - \beta^{n_y}} \log \left( \frac{\exp(z_y)}{\sum_{j=1}^{C} \exp(z_j)} \right)
\end{equation}

In normal semantic segmentation settings, multi-class cross entropy loss is sufficient with \(z\) being predicted logits and \(y\) being the ground truth labels. However, in Occ3D-nuScenes dataset, the class imbalance is extreme. For example, \textit{bicycle} and \textit{motorcycle} class is \( 10^{4} \) times lower in frequency than \textit{drivable space} and \textit{free} class \cite{42dot}. Therefore, a class-balanced loss is applied \cite{cbloss}, where an effective number \(\beta\) is used to reweight the cross entropy loss function, giving more weight to classes with fewer effective samples. Effective number is set to \(\beta = 0.9\). Effective samples \(\beta^{n_y}\) is calculated by the normalized number of voxels representing each class over all ground truth voxels over the entire training dataset. Validity of the class balanced loss is evaluated in the ablation study.


\begin{table*}[thpb]
\caption{Quantitative results of 16 semantic classes on Occ3D-nuScenes validation dataset.}
\label{tab:3D_sem_seg_iou}
\centering
\renewcommand{\arraystretch}{1.5}
\resizebox{1.0\textwidth}{!}{%
\begin{tabular}{lccccccccccccccccc}
\toprule
& \rotatebox{90}{mIoU} & \rotatebox{90}{Barrier} & \rotatebox{90}{Bicycle} & \rotatebox{90}{Bus} & \rotatebox{90}{Car} & \rotatebox{90}{Cons. Veh.} & \rotatebox{90}{Motorcycle} & \rotatebox{90}{Pedestrian} & \rotatebox{90}{Traf. Cone} & \rotatebox{90}{Trailer} & \rotatebox{90}{Truck} & \rotatebox{90}{Driv. Sur.} & \rotatebox{90}{Other Flat} & \rotatebox{90}{Sidewalk} & \rotatebox{90}{Terrain} & \rotatebox{90}{Manmade} & \rotatebox{90}{Vegetation} \\
\midrule
MonoScene \cite{monoscene} (\%) & 6.06 & 7.23 & 4.26 & 4.93 & 9.38 & 5.67 & 3.98 & 3.01 & 5.90 & 4.45 & 7.17 & 14.91 & 6.32 & 7.92 & 7.43 & 1.01 & 7.65 \\
OccFormer \cite{occformer} (\%) & 21.93 & 30.29 & 12.32 & 34.40 & 39.17 & 14.44 & 16.45 & 17.22 & 9.27 & 13.90 & 26.36 & 50.99 & 30.96 & 34.66 & 22.73 & 6.76 & 6.97 \\
BEVFormer \cite{bevformer} (\%) & 23.70 & 36.77 & 11.70 & 29.87 & 38.92 & 10.29 & 22.05 & 16.21 & 14.69 & 26.44 & 23.13 & 48.19 & 33.10 & 29.80 & 17.64 & 19.01 & 13.75 \\
TPVFormer \cite{tpvformer} (\%) & 27.83 & 38.90 & 13.67 & 40.78 & 45.90 & 17.34 & 19.99 & 18.85 & 14.30 & 26.69 & 34.17 & 55.65 & 35.47 & 37.55 & 30.70 & 19.40 & 16.78 \\
CTF-Occ \cite{occ3d} (\%) & 28.53 & 39.33 & 20.56 & 38.29 & 42.24 & 16.93 & 24.52 & 22.72 & 21.05 & 22.98 & 31.11 & 53.33 & 33.84 & 37.98 & 33.23 & 20.79 & 18.0 \\
NVOCC \cite{nvocc} (\%) & 54.19 & 57.98 & 46.40 & 52.36 & 63.07 & 35.68 & 48.81 & 42.98 & 41.75 & 60.82 & 49.56 & 87.29 & 58.29 & 65.93 & 63.30 & 64.28 & 53.76 \\
\textbf{Ours} (\%) & 36.03 & 40.12 & \underline{0.00} & 51.01 & 63.10 & 12.52 & \underline{3.60} & 30.16 & \underline{7.69} & 32.71 & 32.17 & 87.19 & 37.42 & 41.34 & 44.46 & 74.24 & 84.68 \\
\bottomrule
\end{tabular}
}
\end{table*}

\begin{figure*}[thpb]
    \centering
    \parbox{6in}{
        \centering
        \includegraphics[scale=0.21]{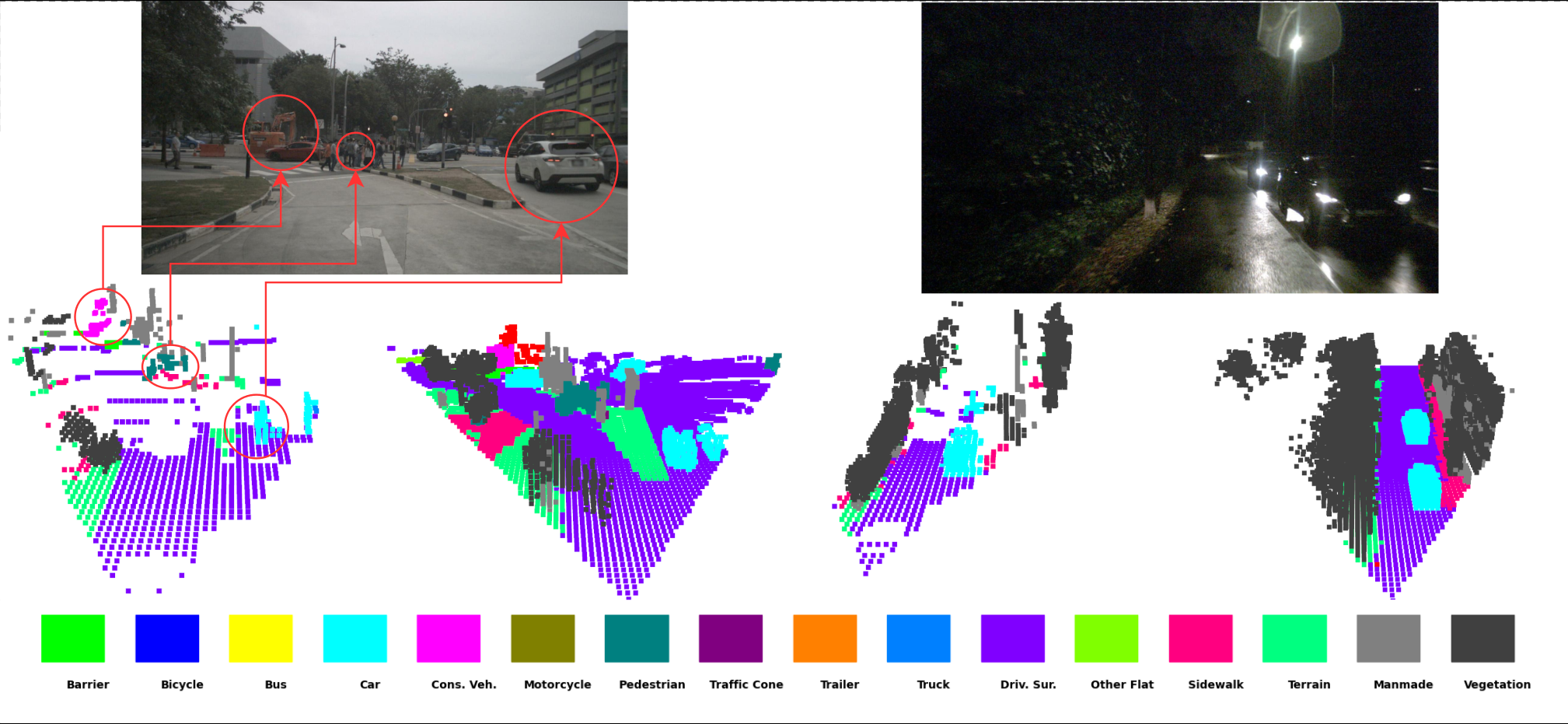}
    }
    \caption{Qualitative predictions on Occ3D-nuScenes validation dataset. Semantic occupancy grid is viewed in 3rd person perspective angle. Each group of images consists of occupancy prediction image on the left, ground truth label on the right, and front camera image on top. }
    \label{quali_pics}
\end{figure*}

\section{Experiments}
\subsection{Experimental Setup}

The training was configured with the Adam optimizer at a learning rate of 1e-4, and a Cosine-Annealing learning rate schedule. The model was trained till convergence, with a batch size of 10. The hardware setup for the training included 1 NVIDIA RTX 4090 GPU and 24GB DDR4 RAM. Training VRAM usage is at 4.3 GB.

Data augmentation is performed on the input and ground truth dataset, which includes image features and LiDAR intensity value perturbation to simulate sensor noise (noise strength of 5\% on normalized values), random translation (-4 to 4 voxel grids) and ground truth voxel masking (10\%).

\subsection{Scene Completion}
We evaluate the 3D scene completion performance using binary voxel completion intersect over union (IoU), precision, recall, and F1 score to be compared with other scene completion methods. Referring to Table \ref{tab:eval_metrics}, \( P \) represents the set of voxels predicted by the model, \( GT \) the set of voxels in the dense occupancy ground truth.

\begin{table}[h]
\caption{Evaluation metrics for 3D scene completion task.}
\label{tab:eval_metrics}
\centering
\begin{tabular}{lccc}
\toprule
Metric & Equation & Better \\
\midrule
IoU & $\frac{|P \cap GT|}{|P \cup GT|}$ & Higher \\
Precision & $\frac{|P \cap GT|}{|P|}$ & Higher \\
Recall & $\frac{|P \cap GT|}{|GT|}$ & Higher \\
F1 Score & $2 \times \frac{\text{Precision} \times \text{Recall}}{\text{Precision} + \text{Recall}}$ & Higher \\
\bottomrule
\end{tabular}
\end{table}

\begin{table}[h]
\caption{Scene Completion evaluation on Occ3D-nuScenes validation set.}
\label{tab:scene_completion}
\centering
\begin{tabular}{lcccc}
\toprule
Model & IoU & Precision & Recall & F1 Score \\
\midrule
TransformerFusion \cite{transformerfusion} & - & 0.375 & 0.591 & 0.453 \\
Atlas \cite{atlas} & - & 0.407 & 0.546 & 0.458 \\
SurroundOcc \cite{surroundocc} & 0.347 & 0.414 & 0.602 & 0.483 \\
MonoScene \cite{monoscene} & 0.342 & - & - & - \\
CTF-Occ \cite{occ3d} & 0.377 & - & - & - \\
\midrule
Ours & \textbf{0.533} & \textbf{0.775} & \textbf{0.632} & \textbf{0.690} \\
\bottomrule
\end{tabular}
\end{table}

We compare these metrics with other scene completion methods on the Occ3D-nuScenes validation dataset. In particular, TransformerFusion \cite{transformerfusion} and Atlas \cite{atlas}, which originally targets the task of 3D reconstruction of indoor scenes from RGB-D images. SurroundOcc \cite{surroundocc}, MonoScene \cite{monoscene} and CTF-Occ \cite{occ3d} focuses on the same outdoor driving scene completion task. Values from these methods are achieved through dense ground truth supervision for fair comparison, as reported by \cite{surroundocc}. Referring to Table \ref{tab:scene_completion}, our proposed method achieves an scene completion IoU of 0.533, outperforming the current state-of-art by 16\%. We also show a large improvement in precision, recall and f1, indicating we densify much more of the sparse input scene while being accurate to the ground truth.

\begin{figure}[thpb]
    \centering
    \parbox{3.3in}{
        \centering
        \includegraphics[scale=0.65]{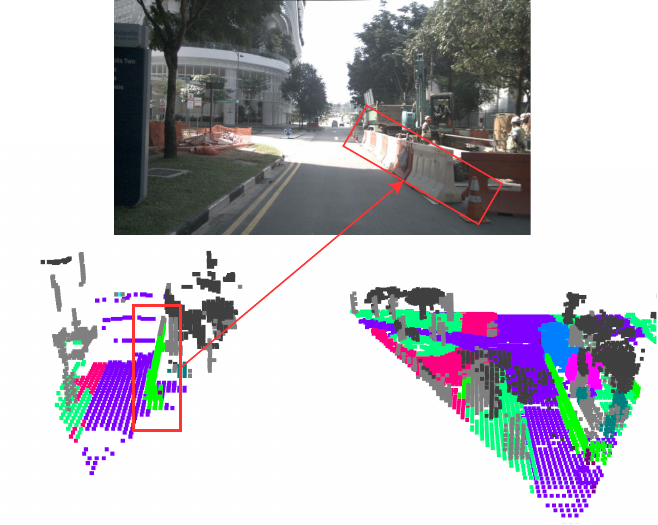} 
    }
    \caption{This example shows the model's ability to predict dynamic obstacles such as temporary road barriers installed for construction on drivable surface.}
    \label{appendex_quali_pics1}
\end{figure}

\begin{figure}[thpb]
    \centering
    \parbox{3in}{
        \centering
        \includegraphics[scale=0.65]{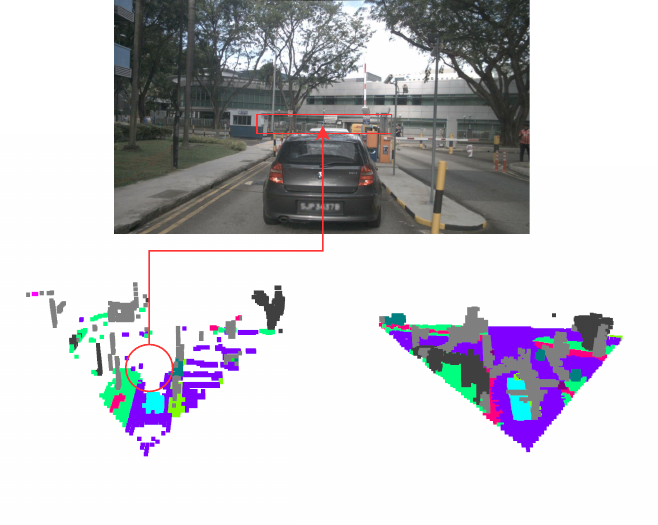} 
    }
    \caption{This example shows the model's inability to hallucinate occluded areas caused by blocking cars. This is the expected performance as the model should not imagine areas without a proper occlusion handling method.}
    \label{appendex_quali_pics2}
\end{figure}

\begin{figure}[thpb]
    \centering
    \parbox{3in}{
        \centering
        \includegraphics[scale=0.65]{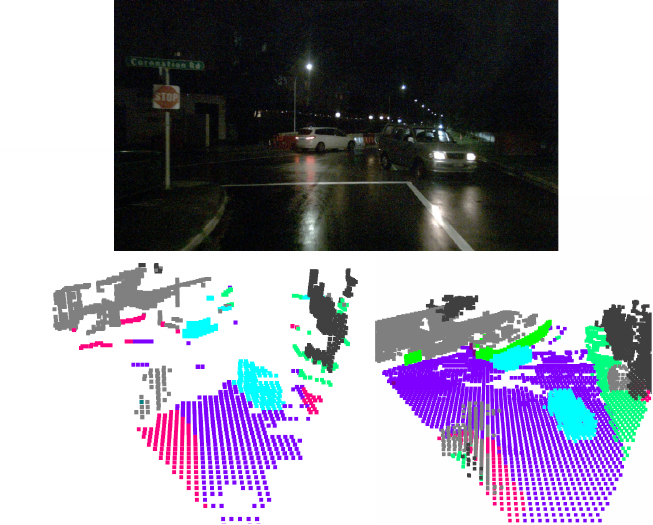} 
    }
    \caption{This example shows a night time intersection with multiple vehicles. The model is able to detect both vehicles with their geometries. However, night time prediction of drivable surface is poor and further distances are empty.}
    \label{appendex_quali_pics3}
\end{figure}

\subsection{Semantic Segmentation}

In semantic segmentation, mIoU (mean Intersection over Union) is calculated using the following formula:
\(\text{mIoU} = \frac{1}{N} \sum_{i=1}^{N} \frac{TP_i}{TP_i + FP_i + FN_i} \), where \(N\) represents the total number of semantic classes, \(TP_i\) denotes True Positives for class \(i\), \(FP_i\) represents False Positives for class \(i\), and \(FN_i\) corresponds to False Negatives for class \(i\). We monitored the semantic segmentation performance through mean Intersection over Union (mIoU) and semantic IoU for each semantic class, and compare it with baseline BEVFormer \cite{bevformer}, several popular architectures \cite{tpvformer, occ3d, occformer} and NVOCC \cite{nvocc}, current state-of-the-art. We do not compare the \textit{void} and \textit{free} class. Referring to Table \ref{tab:3D_sem_seg_iou}, we observe a 8 - 10 \% improvement in mIoU over BEVFormer and CTF-Occ, as well as improvements in most semantic class IoU. Compared to NVOCC, we are competitive in \textit{car}, \textit{drivable surface}, \textit{manmade} and \textit{vegetation}. However, \textit{bicycle}, \textit{motorcycle} and \textit{traffic cones} presents a significant challenge to the model's prediction. These small objects not only appear a lot less in Occ3D-nuScene's dataset due to class imbalance, but also potentially smoothed in the voxelization and convolution processes, erasing or blending their features with its surroundings. Methods to mitigate this problem will be discussed in section VI. 

Figure \ref{quali_pics} shows qualitative inferences of the model. Left group of day-time images demonstrates the model's ability to classify scene objects of varying distance and shapes into distinct semantic classes. Right group of images shows model's fidelity in low light environments: it is able to discern dark colored vehicles at night, as well as predicting obstacles on the left side of the drivable surface. However, the model does show a decrease in maximum prediction distance. More qualitative results are shown in Figure \ref{appendex_quali_pics1}, \ref{appendex_quali_pics2}, \ref{appendex_quali_pics3}.

\subsection{Real-time Inference}
Inference are performed on RTX 4090 with six multi-camera images for the other models at 1600x900 image resolution. Since our model focuses on monocular camera, we use a batch size of 6 for fair comparison. We compare the inference time and GPU memory usage in Table \ref{real_time_inference}, showing a 6 - 10 times increase in inference time, translating to a potential frame rate of 30 FPS and an achievable frame rate of 20 FPS while utilizing only 1.2GB of GPU memory. \\

\begin{table}[h]
\caption{Model time efficiency and GPU memory usage comparison}
\label{real_time_inference}
\centering
\resizebox{.48\textwidth}{!}{%
\begin{tabular}{lcc}
\toprule
Model & Inference Time (s) & Memory Usage (MB)\\
\midrule
BEVformer \cite{bevformer} & 0.31 & 4500 \\
MonoScene \cite{monoscene} & 0.87 & 20300 \\
SurroundOcc \cite{surroundocc} & 0.34 & 5900\\
CTF-Occ \cite{occ3d} & 0.38 & 18000 \\
\midrule
Ours & \textbf{0.03 - 0.05} & \textbf{1200} \\
\bottomrule
\end{tabular}
}
\end{table}

\subsection{Ablation Study}
Referring to Table \ref{table_ablation_study}, we conducted ablation studies by removing the class balanced cross entropy loss and use standard focal loss, removing the squeeze and excite layer, and removing the higher level feature input tensor. "Ours" refers to the current model with all three components included. The results are validated against Occ3D-nuScenes validation dataset. The semantic segmentation mIoU and completion IoU are reported. We observed that removing the class balance loss results in a severe drop in both completion IoU and mIoU. This validates the problem of class imbalance in Occ3D-nuScenes dataset, and class balance loss correctly addresses it by leading the model to be sensitive to all classes in both scene completion and semantic segmentation. The incorporation of Squeeze-and-Excite layers leads to slight improvements in both mIoU and completion IoU. This can be attributed to the use of global average pooling, which enhances the model's adaptability to varying outdoor scenes. Lastly, concatenating higher level image features with the initial \(RGB + Intensity \) features achieved no significant performance improvements. This outcome could suggest that the initial feature set was likely sufficient for the task. It could also indicate the feature extractor, EfficientNetV2 \cite{efficientnetv2}, might have limited capacity to produce meaningful outdoor scene features. 

\begin{table}[h]
\caption{Ablation Study on three core model components.}
\label{table_ablation_study}
\centering
\renewcommand{\arraystretch}{1.1} 
\resizebox{.45\textwidth}{!}{%
\begin{tabular}{lcc}
\toprule
Description & Semseg mIoU & Completion IoU \\
\midrule
Ours & 0.3603 & \textbf{0.5329} \\
Ours w/o CBLoss & 0.3240 & 0.3911 \\
Ours w/o SELayer & 0.3360 & 0.4983 \\
Ours w/o features & \textbf{0.3622} & 0.5110 \\
\bottomrule
\end{tabular}
}
\end{table}

Referring to table \ref{lambda_constant}, using a lambda value of 0.5 yields the best results, with a minor yet notable improvement in Semantic Segmentation mIoU to 0.3555 and a more significant enhancement in Completion IoU to 0.5243. 

\begin{table}[h]
\caption{Lambda constant effect on 3D semantic occupancy prediction.}
\label{lambda_constant}
\centering
\renewcommand{\arraystretch}{1.0} 
\resizebox{0.35\textwidth}{!}{
\begin{tabular}{ccc}
\toprule
$\lambda$ & Semseg mIoU & Completion IoU \\ \midrule
0.3 & 0.3534 & 0.5089 \\
0.5 & \textbf{0.3555} & \textbf{0.5243}\\
0.7 & 0.3466 & 0.5115 \\
\bottomrule
\end{tabular}
}
\end{table}

\section{Conclusions and future work}

In this paper, we presented a Minkowski engine-based sparse convolution model capable of semantic scene completion on diverse outdoor autonomous vehicle driving scenes using LiDAR and a front-view camera. In particular, we demonstrated that the model is able to perform 3D semantic occupancy prediction with significantly fewer computational resources compared to many current methods, achieving real-time inference while maintaining comparable accuracy. More broadly, the sparse convolution approach we propose for AVs offers a pathway to enhance the efficiency of real-time robotics perception.

To address the challenge of detecting and segmenting small objects, incorporating an object detection preprocessing step on the RGB images can be considered. By utilizing a pretrained lightweight object detection module, the 2D bounding boxes can be projected into the 3D space using the same camera-to-LiDAR transformation matrices, subsequently creating 3D ROI (Region of Interest) masks. These masks complement the input sparse tensor into the current network, providing guidance towards small objects during training and improving their segmentation accuracy.

Future work involves several aspects. Firstly, we aim to extend the current model to a multi-view camera setup for 360-degree occupancy prediction of the ego vehicle. Secondly, performing scene completion at distances beyond 20 meters proves challenging due to the sparsity of LiDAR points, specifically when using a 32-beam setup as in the nuScenes dataset \cite{nuscenes}. Incorporating probabilistic techniques to fuse distant camera features and radars into the 3D space may help achieve better accuracy in these regions. Thirdly, employing self-supervision by creating pseudo-dense occupancy ground truth via monocular 2D camera-based depth estimation and semantic segmentation can help expand the training dataset, allowing the current model to become more robust in complex driving environments.

\addtolength{\textheight}{0cm}   





\section*{ACKNOWLEDGMENT}
This project was supported by the EPSRC project RAILS (EP/W011344/1) and the Oxford Robotics Institute's research project RobotCycle. 

\bibliographystyle{IEEEtran}
\bibliography{references}


\end{document}